# Constructing Colloquial Dataset for Persian Sentiment Analysis of Social Microblogs

Mojtaba Mazoochi, Faculty member in ICT Research Institute, Tehran, Iran, mazoochi@itrc.ac.ir

Leila Rabiei, Iran Telecommunication Research Center (ITRC), Tehran, Iran, l.rabiei@itrc.ac.ir

Farzaneh Rahmani, Faculty member in Computer Department, Mehralborz University, Tehran, Iran, rahmani@mehralborz.ac.ir

Zeinab Rajabi[1], Faculty member in Computer Department, Hazrat-e Masoumeh University, Qom, Iran, z.rajabi@hmu.ac.ir

**Abstract:**

**Introduction**: Microblogging websites have massed rich data sources for sentiment analysis and opinion mining. In this regard, sentiment classification has frequently proven inefficient because microblog posts typically lack syntactically consistent terms and representatives since users on these social networks do not like to write lengthy statements. Also, there are some limitations to low-resource languages. The Persian language has exceptional characteristics and demands unique annotated data and models for the sentiment analysis task, which are distinctive from text features within the English dialect.

**Method**: This paper first constructs a user opinion dataset called ITRC-Opinion in a collaborative environment and insource way. Our dataset contains 60,000 informal and colloquial Persian texts from social microblogs such as Twitter and Instagram. Second, this study proposes a new architecture based on the convolutional neural network (CNN) model for more effective sentiment analysis of colloquial text in social microblog posts. The constructed datasets are used to evaluate the presented architecture. Furthermore, some models, such as LSTM, CNN-RNN, BiLSTM, and BiGRU with different word embeddings, including Fasttext, Glove, and Word2vec, investigated our dataset and evaluated the results.

**Results**: The results demonstrate the benefit of our dataset and the proposed model (72% accuracy), displaying meaningful improvement in sentiment classification performance.

[1]Corresponding author

## 1. Introduction

Individuals can now effortlessly and freely express their opinions about any event or entity, such as products, people, services, and others, thanks to the development of social networks. Sentiment analysis is a field of study in which the polarity of opinions on entities, topics, individuals, issues, or events is detected [1, 2]. It is used in several fields, including customer services, internet marketing, and attitude detection.

Many studies have been conducted to enhance the performance of automatic sentiment classification, ranging from light traditional classification methods to more intricate neural network methods. However, little work has been performed on Persian texts. In contrast to English sentiment analysis, Persian sentiment analysis is a novel field of study.

The available sentiment analysis techniques can be classified into lexicon-based and machine-learning approaches. Most approaches developed for the polarity classification in Persian text apply traditional machine learning techniques, which profoundly depend on feature engineering. As a result, extracted features can serve a crucial role in classification satisfaction. Deep learning has recently achieved great success in polarity identification and is considered an acceptable model in various languages. However, the recent performance for sentiment still requires enhancements. Because of the lack of suitable datasets, an intricate syntax structure, and numerous dialects, the Persian language has many weaknesses. In their study, Rajabi and Valavi [3] explain constraints in Persian sentiment analysis. They classified Persian textual constraints into three categories: preprocessing, cultural, and established resource constraints. Deep learning models have recently gained popularity. This rise of interest is because of its successful role in various natural language processing functions. This issue encouraged us to investigate different deep-learning models to enhance the Persian analysis performance.

Sentiment analysis methods require text and lexicon labeled with polarity values. Data with sentiment values play a crucial function in supervised learning methods. In addition, deep learning methods, which have recently shown promising results, necessitate a large amount of labeled data. Furthermore, hybrid approaches merge sentiment lexicon sources with learning methods to significantly increase accuracy.

One challenge in this field is constructing annotated datasets. This study focuses on social networks such as Instagram and Twitter, unlike previous studies such as [4], [5], and [6], which use product review datasets in eCommerce websites for the Persian language. This paper constructs a new opinion dataset for Persian text named ITRC-Opinion. To this end, data are crawled from Instagram and Twitter, and human experts built the dataset. A new deep CNN model using a constructed dataset for Persian texts is proposed to present a new deep learning sentiment model for Persian sentiment analysis. People often use colloquial and spoken terms such as “ واسه، می‌تونم، کتابارو” in opinion texts. The presented model and dataset support informal and colloquial sentences in Persian text. It analyzes posts and comments sent to social microblog media and determines sentiment value at the document level. Pre-trained word embedding for Persian text, such as FastText, Glove, and Word2vec, is applied to the presented model. Furthermore, several other models, such as LSTM, CNN-RNN, BiLSTM, and BiGRU, are examined for a new dataset. The results indicate that, by comparison, the CNN model has the best performance.

The contributions of this article can be summarized as follows:

- Constructing a Persian opinion dataset called ITRC_Opinion[1] in a collaborative environment, focusing on microblogging data and colloquial and informal text, based on crawled data from Instagram and Twitter social networks.

[1] The datasets generated during the current study are available from the corresponding author on reasonable request.

- Proposing a new deep learning architecture for Persian sentiment analysis based on the CNN network and using the constructed dataset, supporting colloquial sentences and informal terms and sentences.

This paper is organized as follows: Section 2 presents a review of relevant studies. In Section 3, the details of constructing the dataset are introduced. Section 4 describes the proposed model. Section 5 reports the evaluation performance and experimental result of our model. Finally, section 6 concludes the paper and makes suggestions for future work.

## 2. Related works

### 2.1. Persian dataset

So far, some sentiment lexicon resources have been created for the Persian language, such as LexiPers Sentiment Lexicon [7], Persian lexicons with polarity labels [8], and Persian Sentiment Analysis and Opinion Mining Lexicon (PreSent) [9, 10]. User opinion datasets are other essential resources for sentiment analysis. They contain sentences/documents with polarity values. In Persian, SentiPers [6], HelloKish [5], and MirasOpinion[4] datasets were developed. Golazizian et al. [11] constructed the first dataset for irony detection in the Persian language. In their study, a Persian dataset for sarcasm detection, called MirasIrony, containing 4339 manually annotated tweets, is created.

The Standard labeled data for Persian sentiment have shortcomings. There are no suitable data resources for analyzing social microblogs yet. The available resources have been mainly created for service/product opinions and e-commerce areas, and social microblogs have been less considered. In addition, the resources have a limited number of records and are ineffective in many applications, such as deep neural networks. In this regard, we need to provide a well-defined sentiment dataset for microblogs in Persian. In the sentiment analysis techniques, there is no gold standard dataset for sentiment classification. This study aimed to fill in this gap and address the challenge.

## 2.2. Persian sentiment methods

Extensive research has been conducted on developing different models for English texts' sentiment analysis. It was first applied in [12] to classify movie reviews into positive and negative classes. They used Support Vector Machine (SVM) and Naïve Bayes classifiers and applied unigram, bag-of-words, and bag-of-concepts [13] as a feature for polarity classification. Later on, many other textual features and learning models have been experimented with by scholars in other studies. Some features applied by researchers were the Part Of Speech (POS) tag, sentiment lexicons and phrases, TF–IDF, syntactic dependency, rules of opinions, and sentiment shifters [1, 2]. The models use a big-sized previously-annotated dataset. Testing and training data are applied for polarity identification.

Some studies have investigated the sentiment analysis of the Persian texts, most of which have focused on traditional classifications. For example, Alimardani and Aghaie [14] [15] used Naïve Bayes, SVM, and Regression logistic algorithms for classification. They use sentiment lexicon for enriching feature engineering. The traditional machine learning approach isolates words into features and applies different feature selection and dimensionality reduction techniques to select the most critical word in the given input. The main methods in this category are SVM, Naive Bayes, and others, which are used to classify the sentiment of a document or a sentence, these models achieved acceptable results in text classification problems. Sentiment and topic modeling simultaneously is an informative task in topic modeling-based sentiment analysis methods [4] [16] [17].

Some studies such as [18] have examined feature engineering for polarity detection. Accordingly, some sentiment features which defined for the polarity classification algorithms of the English text in the latest achievements applied to Persian text. Text features that have been compared include the following: sentiment lexicon, sentiment score of the lexicon, POS tag, TF-IDF, n-gram, word embedding (numeral vectors), and sentiment-specific word embedding.

Recently, active researchers in data science have widely used deep neural networks for analyzing data, including sentiment analysis in different languages. [19]. The key concept in deep learning is that it allows algorithms to realize sentence structure and semantics.

Dashtipour et al. [20] developed two deep learning models, which are CNNs and autoencoders, and applied them to a new dataset. The new dataset exploited in this study was gathered manually and contained Persian film reviews from 2014 to 2016. In [21], Roshanfekr et al. used two deep learning models to classify Persian reviews on digital goods. The data were crawled from the website www.digikala.com, which contains 200761 customer reviews. Dashtipour et al. [22] proposed a combined method for concept-level sentiment analysis in the Persian text merging deep neural networks and linguistic rules to improve sentiment classification. In this method, when a rule is triggered, the method allows sentiments to flow from lexicons to concepts based on symbolic dependency relations. Whenever no pattern is triggered, the method substitutes its subsymbolic counterpart and leverages deep neural networks to classify hotel and product reviews datasets. Bokaee Nezhad et al. [23] proposed a hybrid deep neural network model for the Persian sentiment analysis. According to this model, local features are extracted using CNN, and long-term dependencies are trained by Long Short-Term Memory (LSTM). Zobeidi and Naderan [24] presented a system classifying opinions at the sentence level depend on emotions into two/multiple classes using deep neural network algorithms.

Ghasemi et al. [25] present a bi-lingual deep neural network framework to take advantage of the existing training dataset of English language. They develop bi-lingual word embedding to model sentiment analysis as a transfer learning model, which transfers a model from a resource-rich language to a low-resource one. The bi-lingual sentiment model learns a model on both English and Persian corpora and applies it on Persian text.

Many sentiment analysis models have reached good outcomes for English texts [26], [19], [27, 28]. However, these models may not achieve the same results in Persian text. In addition, some text classification features originating from the available methods for English texts are not effectual for analyzing Persian texts and require further improvements. Also, a deep architecture with acceptable results for English texts may not be efficient for Persian language and require reconstruction. Most studies on the Persian language (e.g., [29], [30], [4]) using deep learning methods have worked on product reviews in eCommerce websites and less on microblogs. Our study focuses on analyzing microblogging social networks.

## 3. Persian sentiment dataset construction

In sentiment analysis, possessing a rich and reliable resource is important. In Persian, some dataset is published, but no one is not appropriate and sufficient for the deep learning model. Unlike the previous study constructing a product review dataset from an eCommerce website, this study focuses on social networks such as Instagram and Twitter. This study creates a new dataset from tweets on Twitter and user comments on Instagram from 2016 to 2019. Figure 1 shows an overview of the construction process of a Persian sentiment analysis dataset.

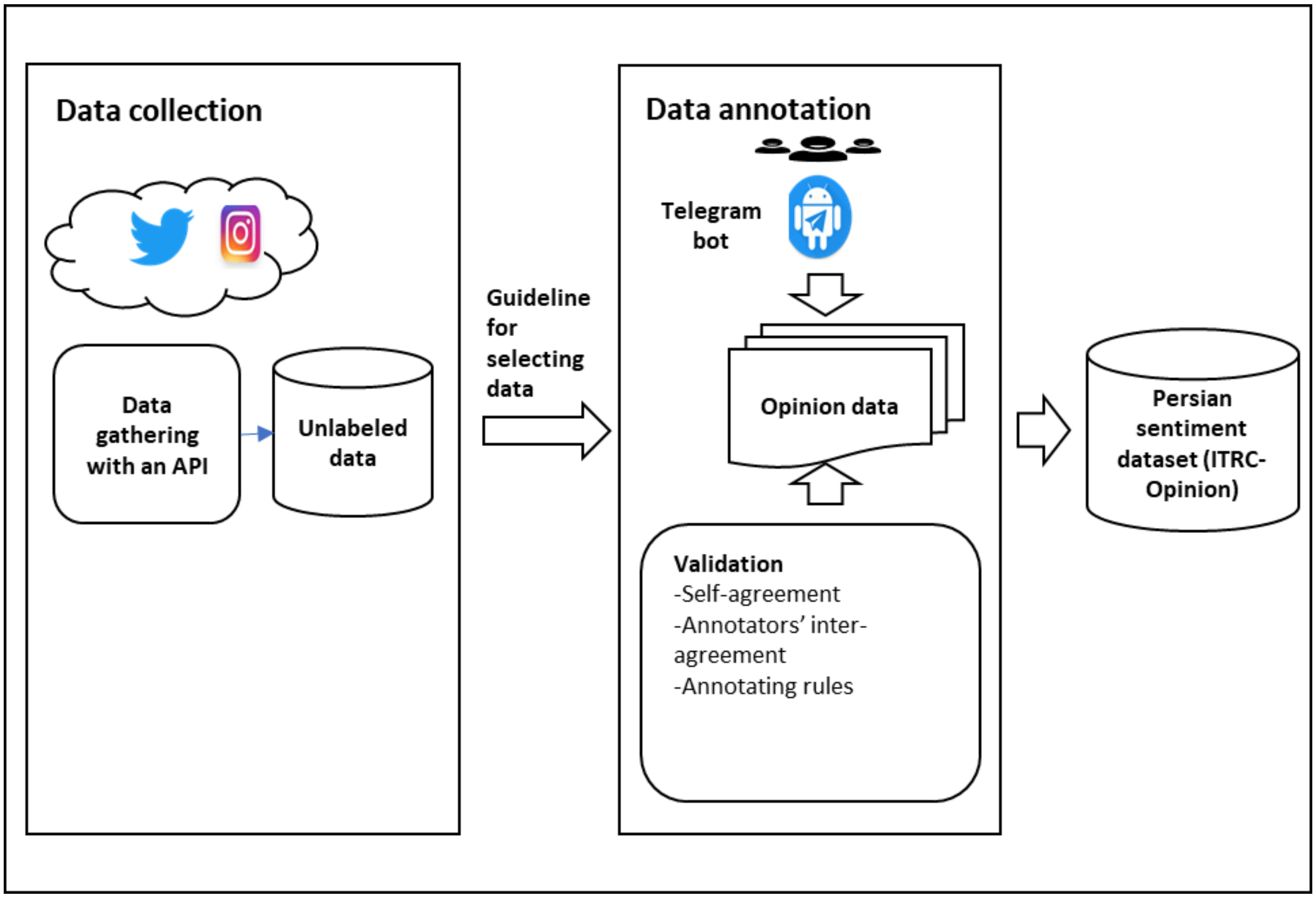

***Figure 1.*** *An overview of the process of constructing the Persian sentiment dataset (ITRC-Opinion)*

Since the importance of opinions in microblogging and social networking services, and existing of different peoples and different writing methods, we crawl data from Twitter and Instagram in this study. In this regard, API tools are designed and created for crawling data from Instagram and Twitter. Finally, the selected data include 60,000 opinions in the Persian language. To cover this issue appropriately, gathered opinions were related to diverse domains.

The remarkable point about these opinions is that they are primarily informal and colloquial and do not obey some grammatical rules. Another case in point was the general data and labeling, which was not limited to specific domains such as politics, economics, and others also extracted from two different social networks and are not limited to opinions in the particular network. Table 1 shows some samples of opinions collected from social networks. Complexity and variability in language are ensured by large amounts of data, random data selection, data crawling from original resources, and data selection from different domains such as politics, economics, and others. The sentences observed by the annotators in the dataset completely acknowledge this issue.

Ten annotators contribute to this corpus, which is trained by reviewing an annotation guideline and the annotation of several sample documents of the corpus. The opinions were manually annotated by native Persian speakers aged 20 to 50.

In addition, all of the annotators are Persian native speakers with proper knowledge and understanding of social media and Persian grammar and some background knowledge of the sentiment analysis. Finally, an experienced annotator reviewed all the annotated documents.

This study selects posts by defined instructions, including the following points: (1) Select posts from multi-domains, as domains are specified; (2)Removing advertisement posts; (3)Selecting posts of identified active users in social networks; (4)Selecting posts with the number of comments more than that of the threshold; and (5)Selecting posts with high likes, and (6)Using comments of specific individuals.

Furthermore, some agreement guidelines are written for standard labeling process and more valid results. They include (1) The advertisement comments have neutral sentiment value; (2)The insult comments have negative sentiment value, (3)The long length comments usually have several sentiment values and require more accuracy to overall sentiment value, (4)The sarcasm comments, which are not insults, can be labeled by positive or negative sentiment, (5)The comments against the homeland are usually negative sentiments, (6)The controversial comments are neutral if addressed to another user, and (7)The emojis can be positive or negative depending on their use in happy or sad situations. For example, the laughing emoji expresses a person's feelings of happiness and indicates positive sentiment; the sentiment value is directly identified by the number of laughing emojis in the tweet. In other words, three laughing emojis would have a much greater sentiment value than one laughing emoji.

***Table 1**. Samples of different types of comments*

| | |
|---|---|
| **Advertisement comment:** | |
| Visit my page if you like to have a minimalist digital painting with your own photo.☺ | اگه دوست دارین نقاشی دیجیتال به سبک مینیمال، با عکس خودتون داشته باشین، به پیج من سر بزنید.☺ |
| **Sarcasm comment:** | |
| It doesn't matter; we didn't used to go out much before either 😂😂😂😂<br>What a steady turkey it is! Its price is gonna increase again tomorrow | همچین فرقی نداره ما قبلن هم بیرون زیاد میرفتیم😂😂😂😂<br>اخی بوقلمون چه سینه سپری کرده از فردا گرون میشه باز |
| **Positive comment:** | |
| That's very kind of you; thanks buddy🌹🌹👑 | خیلی بزرگی دمت گرم🌹🌹👑 |
| **Negative comment:** | |
| May peace be upon you, and his soul rest in peace.❤❤❤ | خدا بهتون صبر بده روحشون شاد❤❤❤ |

In this study, a Telegram bot was specifically created for contributors to assign a polarity score by humans. The humans read the post and then label the polarity of comments at the document level with a Telegram bot. The polarity score to comments ranges from the set -1 to 1, indicating the sentiment orientation of the comment by -1 as the most negative, +1 as the most positive, and '0' indicating the sentence's polarity neutral. Figure 2 depicts the distribution of the dataset's labeled positive, negative, and neutral sentiment classes.

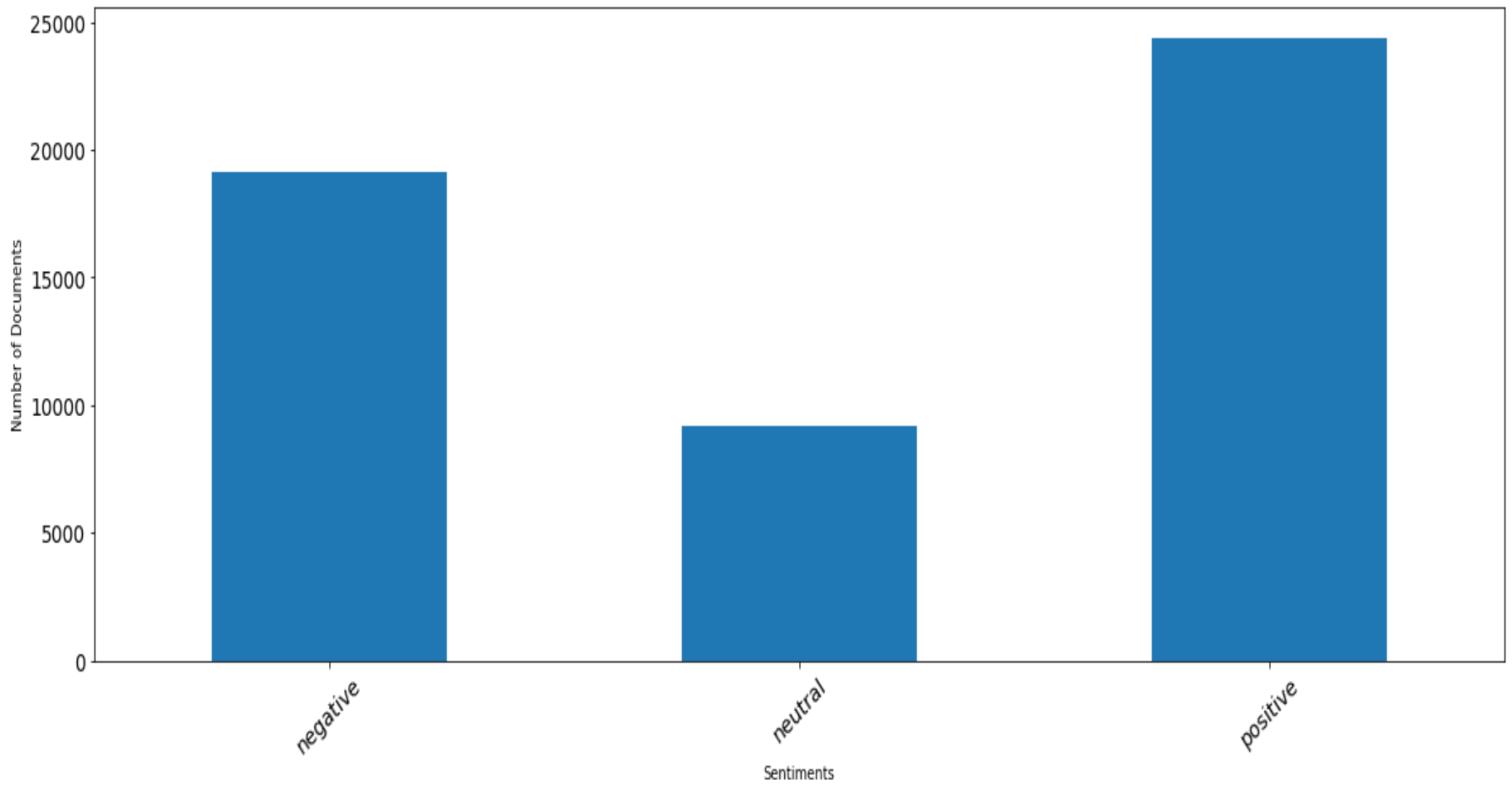

***Figure 2.*** *Distribution of positive, negative, and neutral classes in the dataset*

Figure 3 shows the relationship between the length of each comment and the sentiment label assigned to the samples. As it can be noticed, on average, longer comments have a negative sentiment label. Figure 4 presents the distribution of the length of comments in the ITRC-Opinion dataset. As it can be noticed, comments with a length of less than 7 have the highest number of labeled comments. Figure 5 shows the number of emojis in the comments. As it can be noticed, more than 60% of the data did not have emojis, about 10% had one emoji, and very few texts had a large number of emojis.

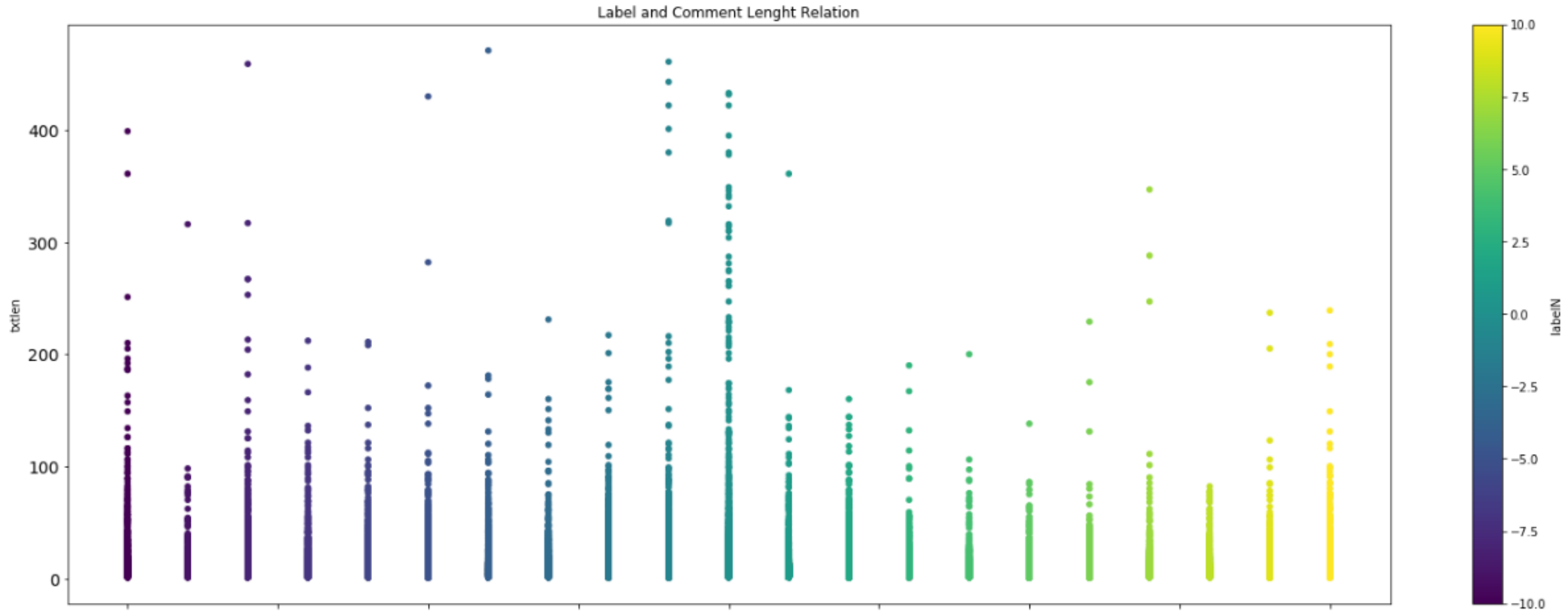

***Figure 3.*** *The relationship between the length of each comment and the sentiment label assigned to the samples. Txtlen shows comments' length. LablelN shows the sentiment label. The most negative sentiment is determined by dark blue, and the most positive one is determined by yellow.*

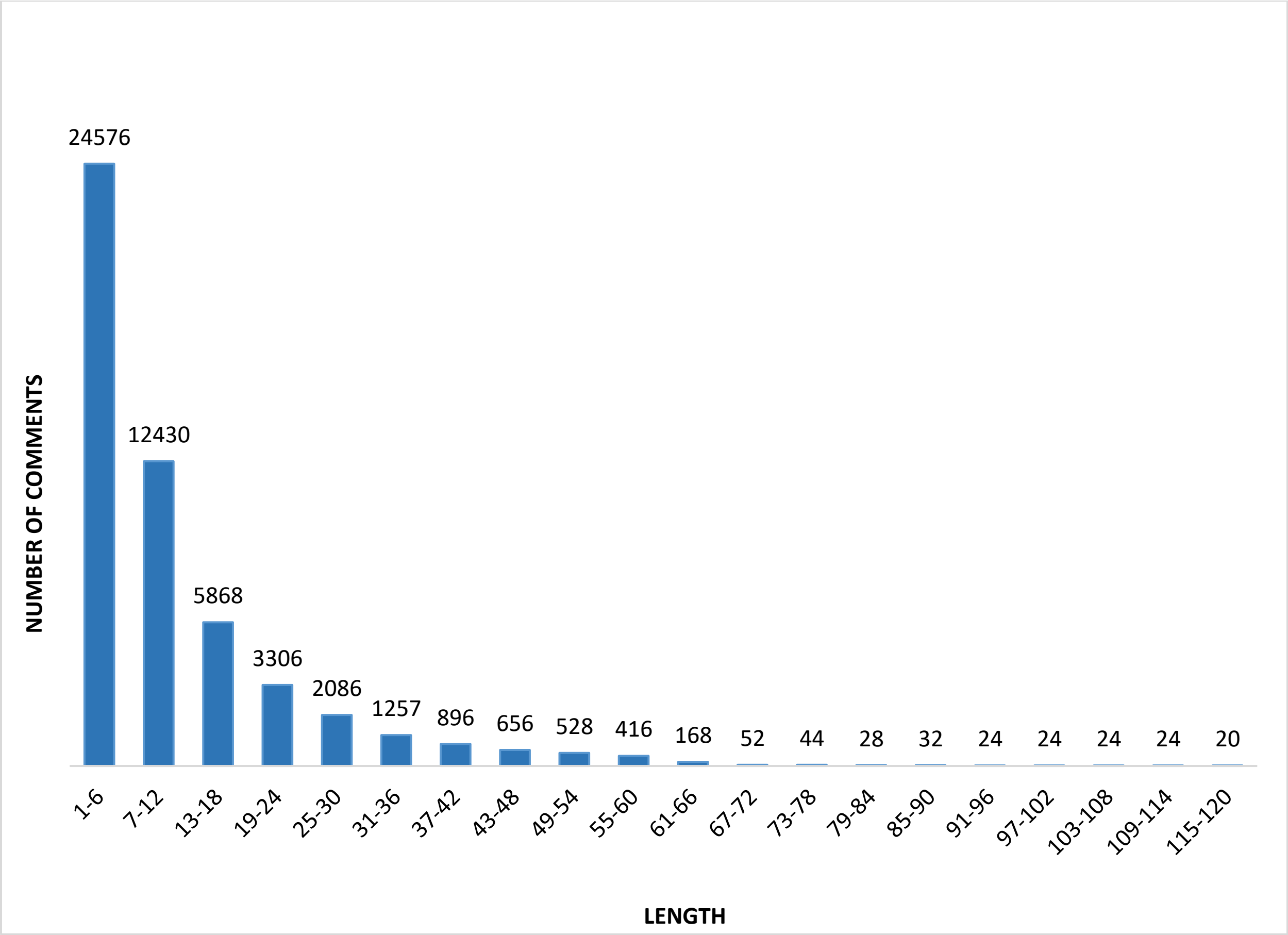

***Figure 4.*** *Comments' length distribution in the dataset.*

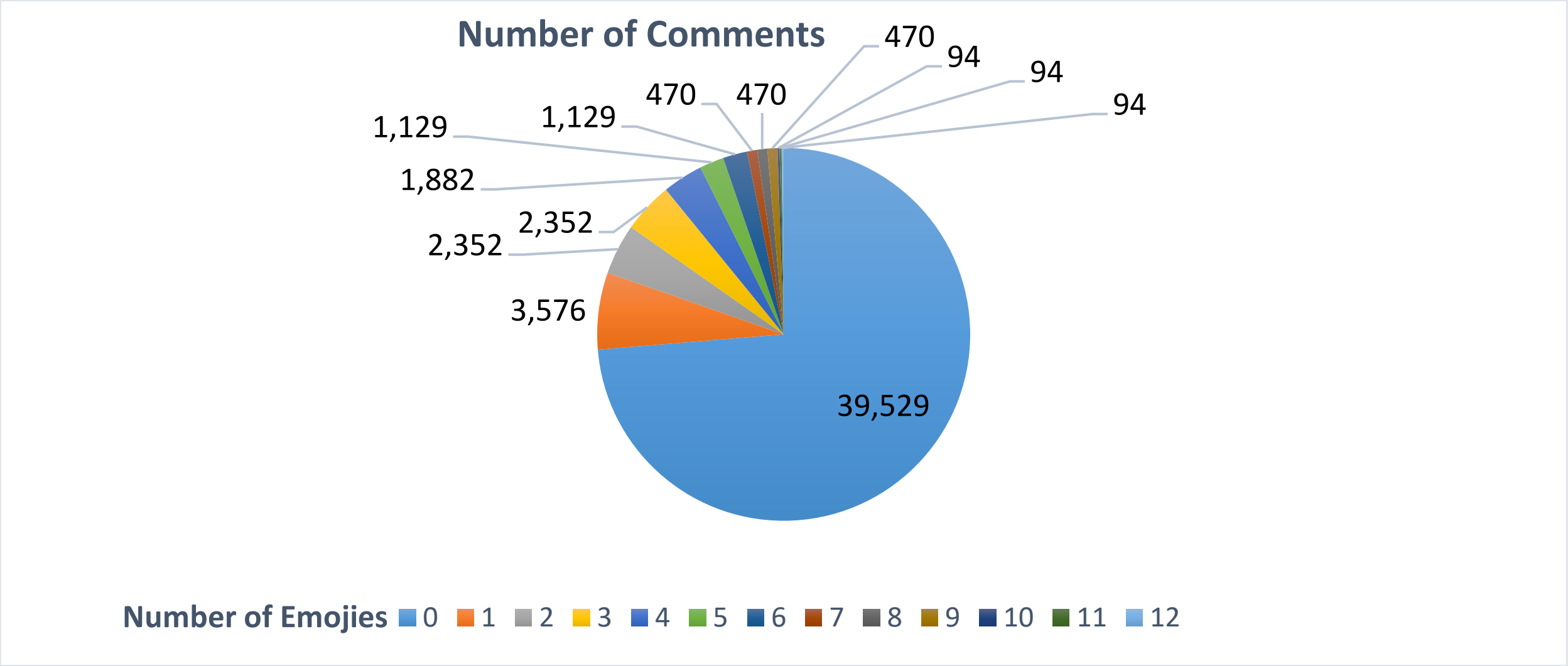

***Figure 5.*** *Relation between number of comments and number of emojis in the dataset*

For verification and validation of users' assigned polarity labels, we used two methods for measuring these purposes, the first one is the inter_annotator agreement, which calculates the agreement among the annotators. Some of the well-known measures for the calculation of inter-annotator agreement are Fleiss's K, Cohen's Kappa, Cronbach's Alpha, and Krippendorff's Alpha [27]. Fleiss' kappa is calculated for a group of annotators. It is considered a proper measure because the values of polarities assigned to sentences are of the nominal type. In Fleiss's kappa formula, three categories have been considered: positive, neutral, and negative. Data labeling is performed in two rounds. In the first round, data with no agreement on the labels assigned by the annotators are separated. In the second round, the disputed sentences are labeled by the annotators once more. In this round, if the majority vote is obtained, the most agreed-upon label is selected, and if there is no agreement, the sentences are removed in some cases. The 67% inter-agreement denotes substantial agreement among the annotators.

The second method is self-agreement, which calculating the agreement between assigned labels to the same comment by each user individually. The result of these methods is shown in Table 2.

***Table 2***. *Validation results*

| Method | Agreement |
| --- | --- |
| Self-agreement | 89% |
| Annotators' inter-agreement | 67% |

## 4. The proposed model

In this section, we explain the main components of the proposed model. Data are preprocessed after the opinion dataset is constructed. In this regard, textual data are cleaned and normalized by preprocessing tools. Also, the unwanted symbols and tokens are removed from the text. In the following steps, data are prepared for the learning process.

### 4.1. Data preprocessing

In this part, we normalize the comments and eliminate unwanted symbols and noisy words. This part aims to maximize the number of lexicons whose embedding can be gained in the pre-trained word embedding coding. The preprocessing steps we follow for normalizing and cleaning Persian comments are as follows:

- The emojis included in the opinions are replaced with their names since the word embedding model does not include vector representation for emojis. To do so, we organized an inventory for assigning emotion icons to their interrelated names. Using this method, the emotions expressed by these emojis are not ignored.
- Date-time, numbers (Persian/English), URL addresses, and unique social media tokens such as @ and # are removed from the text.

- Usually, a group of emoticons appears in a text without space between them. There are no embedding words for this combination. In this condition, tokenizing and normalization are performed.
- The words are normalized, and unwanted punctuation marks and symbols are removed.
- Repetitions and elongations are removed, and a single occurrence is used instead.

Efforts are made to preserve information in the preprocessing steps. For example, emojis included in the opinions are not removed; however, they are replaced with their names. Due to the shortness of the text, the use of abbreviations, and the dynamic nature of language, some new words may be introduced, and these words are preserved and not removed. Moreover, the constructed word embedding based on Fasttext is trained on colloquial Persian text and social network data containing emerging words.

### 4.2. Word embedding

Each comment text contains a variable-length set of words. In the preliminary step, text words are substituted by their corresponding word embedding vector from our produced word embedding on millions of Twitter and Instagram comments. Each comment is represented by a 2D vector of dimension $n \times d$ where n is the number of words in the tweet or Instagram comment and d is the dimension length of the word's vector representation. We use the vector dimension 100 for embedding. To ensure that all comments are of the same fixed size, padding each tweet's representation by zeros.

We train embedding vectors with Word2vec[31], Glove[32], and FastText[33] model, as a result, FastText is the most influential and effective recent model. It represents each word as a bag-of-character n-gram. Representations for character n-grams, once they are learned, can be combined (via simple summation) to represent out-of-vocabulary (OOV) words.

### 4.3. A CNN-based model for sentiment classification

Convolutional neural networks are one of the most important deep learning algorithms and are well-known for their acceptable performance in classification problems. CNN's have many usages in image processing, and nowadays, they are used in the field of NLP as well. The model has been widely used in different sentiment analysis frameworks in the English language and showed promising results[34]. Also, some studies (e.g., [25], [22], [35]) in Persian use the CNN model and achieve acceptable outcomes. Furthermore, CNN mainly focuses on extracting the local features of the text, the fact that attention mechanisms are concerned with information in the context. Convolutional layers only use special connections from the previous layer; local neurons are connected to the subsequent layer neurons. In this research, our proposed model applies the CNN network model, as presented in Figure 6. A CNN architecture is initially developed in this study to predict the polarity of the texts. After texts are represented by word embedding, described in the previous section, we design three main layers, including convolutional, global max pooling, and fully connected for sentiment classification. This study employs an adam optimizer for deep neural network optimization algorithm and achieves the best results by five epochs. The sentiment analysis code and sample dataset are publicly available at https://github.com/zekavat-ITRC/sentiment-Analysis/tree/main.

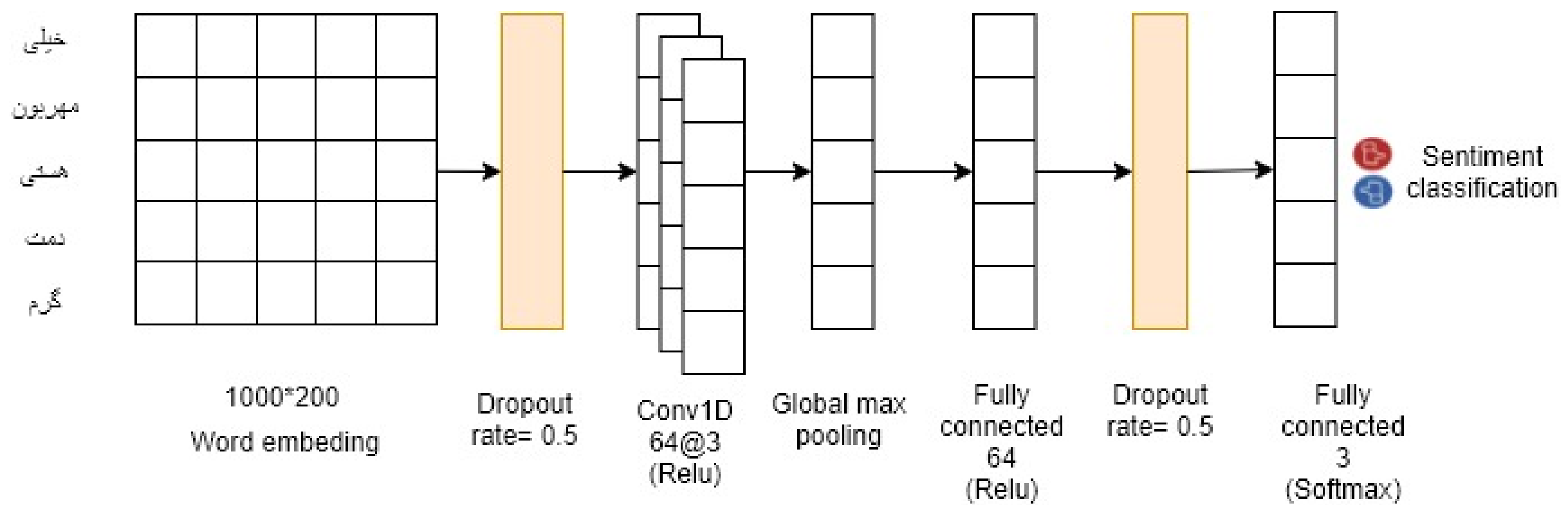

***Figure 6**. Presented convolutional neural network (CNN) architecture for sentiment analysis of Persian text*

## 5. Experimental results

A subset of the dataset, approximately 90%, was used to train the neural network. The remaining data, about 10%, was used to test and validate the learned model's performance: 6% for testing and 4% for validation. There is a range of labels in the dataset from -1 to +1. In the first step, the corpus was preprocessed using the mentioned preprocessing methods, and then FastText was applied to transform each word into 100-dimension numerical vectors. To automatically learn features for polarity detection from training data, different neural network models are proposed. LSTM, CNN, CNN-RNN, BiLSTM, and BiGRU models have been used in multiple experiments for classification.

We design five deep learning models, LSTM, CNN-LSTM, Bidirectional LSTM, Bidirectional GRU, and CNN. First, we train all the models by various hyper-parameters. Then, we choose the most suitable model that yields the highest accuracy.

### 5.1. Hyperparameters analysis

Various experiments were performed on deep neural network models such as GRU, CNN-RNN, CNN, LSTM, and Bidirectional GRU. In each model, we gain the best results in reported hyperparameters. In each experiment of CNN, we changed hyper-parameter values to get the best of them, and finally, the best hyper-parameters are detailed in Table 3. The most suitable hyper-parameters found after many experiments for the LSTM architecture are reported in Table 4. The best hyper-parameter values of CNN-RNN for this task are shown in Table 5. The best hyperparameter values of the Bidirectional GRU architecture for this task are described in Table 6. The best hyperparameter values of the Bidirectional LSTM architecture for this task are detailed in Table 7. As can be seen from Table 8, CNN-based architecture has achieved an acceptable result among all the mentioned models in terms of F1 score and accuracy metric value.

The best accuracy is 72% in the CNN-based model, which is better than other models. In the presented architecture, we had precision=0.66, recall=0.79, and F1-score=0.71 in the best case. Overall, the CNN-based model showed better performance in sentiment analysis of Persian colloquial text in the microblog. It performs appropriately since social network users use short texts, emojis, and abbreviations in everyday texts. We should rely on sentiment keywords, emojis, and remaining words to detect the sentiment of posts. LSTM/BiLSTM methods focus on the sequence of text and achieve high performance when we have long and complete sentences and a formal text. However, the CNN algorithm is defined by its potential to extract local features (n-grams) from the data. Figure 7 shows the accuracy/loss curves of the proposed model for training and testing data. As shown, the proposed model has the best performance in epoch 5. Figure 8 shows the time per epoch(minutes) of different deep learning models with different word embedding. As shown, the CNN-based model with Glove word embedding takes less time than other models. After Glove, the CNN-based model with Word2vec and Fasttext word embedding took less time, respectively.

*Table 3. Hyperparameter settings used in the CNN-based model.*

| **Hyperparameters** | **Value** |
|---|---|
| Number of layers | 7 |
| Number of filters | 128 |
| Filter size | 3 |
| Dropouts rate | 0.5 |
| Number of epochs | 5 |
| Batch size | 8 |

*1*

*Table 4. Hyperparameter settings used in the LSTM-based model.*

| **Hyperparameters** | **Value** |
|---|---|
| Number of layers | 4 |
| LSTM hidden state dimension | [128,64] |
| Dropout rate | [0.4,0.4] |
| Learning rate | 0.001 |
| Number of epochs | 5 |
| Batch size | 128 |

*Table 5 Hyperparameter settings used in the CNN-RNN-based model.*

| **Hyperparameters** | **Value** |
|---|---|
| Number of layers | 10 |
| CNN hyper-parameters | Num of filters: 64, Filter Size: 3 |
| RNN parameters | Cells: GRU, Num of states: 64 |
| Dropout rate | [0.2,0.3,0.5] |
| Learning rate | 0.001 |
| Number of epochs | 5 |
| Batch size | 256 |

***Table 6.*** *Hyperparameter settings used in the GRU-based model.*

| **Hyperparameters** | **Value** |
|---|---|
| Number of layers | 5 |
| GRU hidden state dimension | 64 |
| Spatial dropout rate | 0.3 |
| Learning rate | 0.001 |
| Number of epochs | 6 |
| Batch size | 256 |

***Table 7****. Hyperparameter settings used in the Bi-LSTM-based model*

| **Hyper Parameters** | **Value** |
|---|---|
| Number of layers | 6 |
| LSTM hidden state dimension | 40 |
| Dropout rate | 0.4 |
| Learning rate | 0.008 |
| Number of epochs | 5 |
| Batch size | 256 |

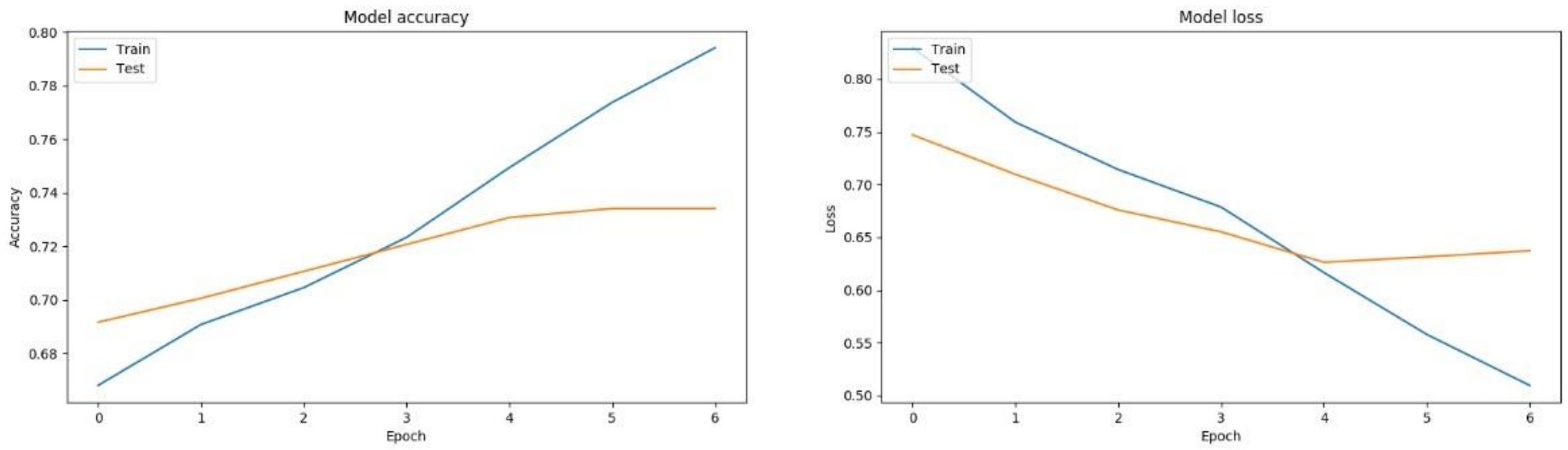

***Figure 7.*** *Loss and accuracy curves of proposed model(training and testing data)*

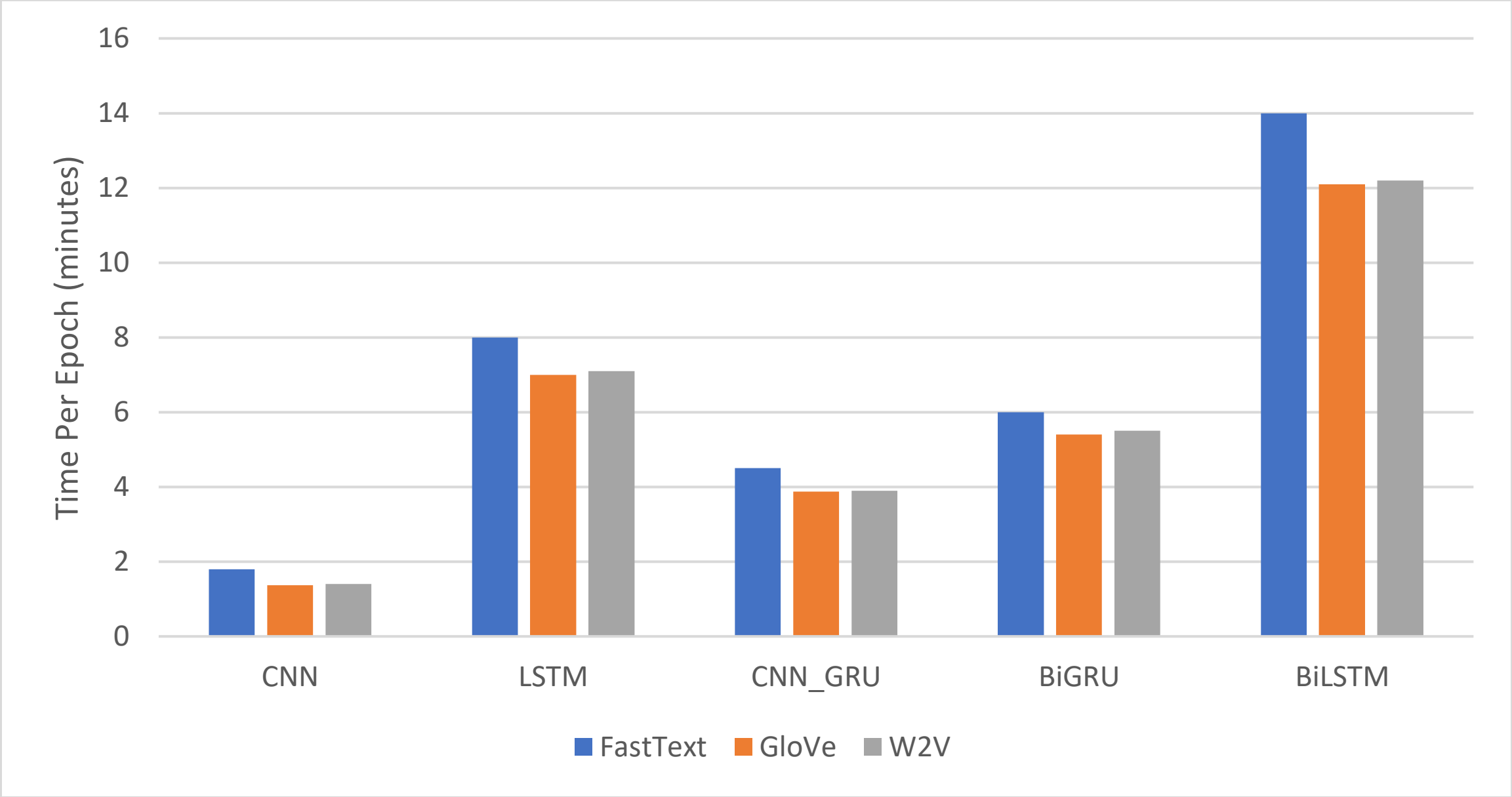

***Figure 8.*** *Time per epoch (minutes) of different deep learning models with different word embeddings*

***Table 8.*** *Comparing the outcomes of the CNN-based sentiment analysis model with the other deep neural models with different word embedding*

| **Model** | | **FastText** | **Glove** | **Word2Vec** |
|---|---|---|---|---|
| **CNN** | **F1-score** | 0.65 | 0.62 | 0.623 |
| | **Accuracy** | **0.72** | 0.664 | 0.659 |
| **LSTM** | **F1-score** | 0.62 | 0.631 | 0.60 |
| | **Accuracy** | 0.66 | 0.65 | 0.65 |
| **CNN-GRU** | **F1-score** | 0.626 | 0.60 | 0.587 |
| | **Accuracy** | 0.674 | 0.658 | 0.659 |
| **Bidirectional GRU** | **F1-score** | 0.613 | 0.629 | 0.62 |
| | **Accuracy** | 0.67 | 0.66 | 0.647 |
| **Bidirectional LSTM** | **F1-score** | 0.628 | 0.605 | 0.61 |
| | **Accuracy** | 0.662 | 0.65 | 0.654 |

The emergence of COVID-19 in Wuhan, China, in December 2019 had significant effects on language. During the COVID-19 pandemic, new words, phrases, and colloquial literature became popular in people's conversations. We select short and slang comments on COVID-19 topics from social networks and evaluate the model's behavior about these sentences. Table 9 shows sample sentences and sentiment predictions of the proposed model. As observed in the examples, however short, incomplete, colloquial abbreviations, interrupted words, colloquial phrases, and new vocabulary are used in sentences, our model predicts the correct sentiment value. The real test cases demonstrate that our model exhibits fewer errors and performs well in colloquial text and emerging literature.

**Table 9. Real test cases**

| **Post text (English)** | **Post text (Persian)** | **Sentiment value** |
|---|---|---|
| Under this condition, there is no hope for future, so I don't think about my goals 😢😂 | با این اوضاع کرونا هیچ امیدی به ایندم ندارم پس دیه به هدفم فک نمیکنم 😢😂 | -1 |
| Be affected by Coronavirus, get sick, and feel relaxed | کرونا بگیری سیک شی راحت شیم | -1 |
| God willing, everyone around the world will be far away from the scourge of Corona, and the sick would get better soon | انشا... که همه تو هر نقطه از کره زمین از بلای کرونا به دور باشه و مریضا زود خوب بشن | +1 |
| I wish this rain would wash and take away Coronavirus. | کاش این بارون کرونا رو هم می‌شست و با خودش می‌برد | +1 |

The advantages of the created dataset compared to other data are the diversity of data and non-domain-specific data. Because our model is created on the microblog and other models implemented in Persian are domain-specific and more products reviewed, a proper comparison cannot be made with other Persian works.

## 6. Conclusion

Analyzing social network data such as Instagram and Twitter is very important. When many annotated data are not available for learning, deep learning methods may not work well. The current study constructed a Persian opinion dataset using posts and comments on microblogs. The research attempted to provide a resource for low-resource language and facilitated sentiment analysis of Persian text. The new dataset applied in this work was gathered by our designed system and included social reviews in the Persian language from 2016 to 2019. The verification and validation of the ITRC-opinion dataset was performed in a standard manner.

In this study, we demonstrated the results of using deep neural network models on the sentiment analysis of Persian comments. Our models only rely on pretrained word vector representations (Embedding). Although the Persian language is very complex and the used model is simple, the CNN-based model achieves significant improvement in F1-score and accuracy compared to other models. The reason for this success is that CNN is defined by the ability to extract local features (n-grams) from the data. The proposed model compared with other models such as LSTM, CNN-RNN, BiLSTM, and BiGRU, and the achieved results illustrate that our CNN-based model performs relatively well.

One of the challenges in deep learning is the emergence of the heavy architecture of some models despite good accuracy. Heavy models do not operate with acceptable performance in practice, and as presented in this article, light models with more appropriate data achieve good results.

The experimental results illustrated that proposed model gained relatively good achievement. According to the results, our model performed well in the sentiment classification of social network text. Accordingly, it is suggested to grow our model with sentiment-specific word embedding for Persian texts in future studies. Sentiment-specific word embedding for Persian text improves the accuracy of the model more than general word embedding and represents polarity contrast better.

**Conflicts of interests**

The authors did not receive support from any organization for the submitted work.

## Data availability

The datasets generated during the current study are available from the corresponding author on reasonable request.

## References

1. Liu, B., *Sentiment analysis and opinion mining*. Synthesis lectures on human language technologies, 2012. **5**(1): p. 1-167.
2. Liu, B., *Sentiment Analysis and Subjectivity*. Handbook of natural language processing, 2010. **2**: p. 627-666.
3. Rajabi, Z. and M. Valavi, *A Survey on Sentiment Analysis in Persian: a Comprehensive System Perspective Covering Challenges and Advances in Resources and Methods*. Cognitive Computation, 2021: p. 1-21.
4. Ashrafi Asli, S.A., B. Sabeti, Z. Majdabadi, P. Golazizian, and O. Momenzadeh. *Optimizing annotation effort using active learning strategies: a sentiment analysis case study in persian*. in *Proceedings of The 12th Language Resources and Evaluation Conference*. 2020.
5. Moradi Mehdi, Khosravizade Parvane, and V. Bahram, *Constructing tagged corpora with a web approach as a corpus*, in *2th symposium on computational Linguistics*. 2012.
6. Hosseini, P., A.A. Ramaki, H. Maleki, M. Anvari, and S.A. Mirroshandel, *SentiPers: a sentiment analysis corpus for Persian*. arXiv preprint arXiv:1801.07737, 2018.
7. Sabeti, B., P. Hosseini, G. Ghassem-Sani, and S.A. Mirroshandel. *LexiPers: An ontology based sentiment lexicon for Persian*. in *GCAI*. 2016.
8. Dehdarbehbahani, I., A. Shakery, and H. Faili, *Semi-supervised word polarity identification in resource-lean languages*. Neural Networks, 2014. **58**: p. 50-59.
9. Dashtipour, K., A. Hussain, Q. Zhou, A. Gelbukh, A.Y. Hawalah, and E. Cambria. *PerSent: a freely available Persian sentiment lexicon*. in *International Conference on Brain Inspired Cognitive Systems*. 2016. Springer.
10. Dashtipour, K., A. Raza, A. Gelbukh, R. Zhang, E. Cambria, and A. Hussain. *PerSent 2.0: Persian Sentiment Lexicon Enriched with Domain-Specific Words*. in *International Conference on Brain Inspired Cognitive Systems*. 2019. Springer.
11. Golazizian, P., B. Sabeti, S.A.A. Asli, Z. Majdabadi, and O. Momenzadeh. *Irony Detection in Persian Language: A Transfer Learning Approach Using Emoji Prediction*. in *Proceedings of The 12th Language Resources and Evaluation Conference*. 2020.
12. Pang, B., L. Lee, and S. Vaithyanathan. *Thumbs up?: sentiment classification using machine learning techniques*. in *Proceedings of the ACL-02 conference on Empirical methods in natural language processing-Volume 10*. 2002. Association for Computational Linguistics.
13. Rajabi, Z., M.R. Valavi, and M. Hourali, *A Context-based Disambiguation Model for Sentiment Concepts Using a Bag-of-concepts Approach*. Cognitive Computation, 2020: p. 1-14.
14. Alimardani, S. and A. Aghaie, *Opinion mining in Persian language using supervised algorithms*. Journal of Information Systems and Telecommunication (JIST), 2015: p. 135-141.

15. Alimardani, S. and A. Aghaie, *Opinion mining in Persian language using supervised algorithms and sentiment lexicon.* Journal of Information Technology Management, 2015. **7**: p. 345-362.
16. Shams, M., A. Shakery, and H. Faili. *A non-parametric LDA-based induction method for sentiment analysis*. in *The 16th CSI International Symposium on Artificial Intelligence and Signal Processing (AISP 2012)*. 2012. IEEE.
17. Vaziripour, E., C. Giraud-Carrier, and D. Zappala. *Analyzing the political sentiment of Tweets in Farsi*. in *Tenth International AAAI Conference on Web and Social Media*. 2016.
18. Asgarian, E., M. Kahani, and S. Sharifi, *The impact of sentiment features on the sentiment polarity classification in Persian reviews.* Cognitive Computation, 2018. **10**(1): p. 117-135.
19. Fu, X., J. Yang, J. Li, M. Fang, and H. Wang, *Lexicon-enhanced LSTM with attention for general sentiment analysis.* IEEE Access, 2018. **6**: p. 71884-71891.
20. Dashtipour, K., M. Gogate, A. Adeel, C. Ieracitano, H. Larijani, and A. Hussain. *Exploiting deep learning for persian sentiment analysis*. in *International Conference on Brain Inspired Cognitive Systems*. 2018. Springer.
21. Roshanfekr, B., S. Khadivi, and M. Rahmati. *Sentiment analysis using deep learning on Persian texts*. in *2017 Iranian Conference on Electrical Engineering (ICEE)*. 2017. IEEE.
22. Dashtipour, K., M. Gogate, J. Li, F. Jiang, B. Kong, and A. Hussain, *A hybrid Persian sentiment analysis framework: Integrating dependency grammar based rules and deep neural networks.* Neurocomputing, 2020. **380**: p. 1-10.
23. Nezhad, Z.B. and M.A. Deihimi, *A combined deep learning model for persian sentiment analysis.* IIUM Engineering Journal, 2019. **20**(1): p. 129-139.
24. Zobeidi, S., M. Naderan, and S.E. Alavi, *Opinion mining in Persian language using a hybrid feature extraction approach based on convolutional neural network.* Multimedia Tools and Applications, 2019. **78**(22): p. 32357-32378.
25. Ghasemi, R., S.A. Ashrafi Asli, and S. Momtazi, *Deep Persian sentiment analysis: Cross-lingual training for low-resource languages.* Journal of Information Science, 2020: p. 0165551520962781.
26. Teng, Z., D.T. Vo, and Y. Zhang. *Context-sensitive lexicon features for neural sentiment analysis*. in *Proceedings of the 2016 conference on empirical methods in natural language processing*. 2016.
27. Zhang, L., S. Wang, and B. Liu, *Deep learning for sentiment analysis: A survey.* Wiley Interdisciplinary Reviews: Data Mining and Knowledge Discovery, 2018. **8**(4): p. e1253.
28. Young, T., D. Hazarika, S. Poria, and E. Cambria, *Recent trends in deep learning based natural language processing.* ieee Computational intelligenCe magazine, 2018. **13**(3): p. 55-75.
29. Dastgheib, M.B., S. Koleini, and F. Rasti, *The application of deep learning in Persian documents sentiment analysis.* International Journal of Information Science and Management (IJISM), 2020. **18**(1): p. 1-15.
30. Bagheri, A., *Integrating word status for joint detection of sentiment and aspect in reviews.* Journal of Information Science, 2019. **45**(6): p. 736-755.
31. Mikolov, T., K. Chen, G. Corrado, and J. Dean, *Efficient estimation of word representations in vector space.* arXiv preprint arXiv:1301.3781, 2013.
32. Pennington, J., R. Socher, and C.D. Manning. *Glove: Global vectors for word representation*. in *Proceedings of the 2014 conference on empirical methods in natural language processing (EMNLP)*. 2014.
33. Athiwaratkun, B., A.G. Wilson, and A. Anandkumar, *Probabilistic fasttext for multi-sense word embeddings.* arXiv preprint arXiv:1806.02901, 2018.
34. Liao, S., J. Wang, R. Yu, K. Sato, and Z. Cheng, *CNN for situations understanding based on sentiment analysis of twitter data.* Procedia computer science, 2017. **111**: p. 376-381.
35. Vazan, M. and J. Razmara, *Jointly Modeling Aspect and Polarity for Aspect-based Sentiment Analysis in Persian Reviews.* arXiv preprint arXiv:2109.07680, 2021.